\documentclass[letterpaper, 10 pt, conference]{ieeeconf}  
\IEEEoverridecommandlockouts                              
\usepackage{graphicx} 
\usepackage{epsfig}
\usepackage{mathptmx} 
\usepackage{times} 
\usepackage{amsmath} 
\usepackage{amssymb} 
\usepackage{hyperref}
\usepackage{url}

\usepackage{xcolor}
\usepackage{layouts}

\usepackage{comment}

\title{\LARGE \bf Locomotion and Control of a Friction-Driven Tripedal Robot}

\author{Mark Hermes$^{1}$, Taylor McLaughlin$^{2}$, Mitul Luhar$^{1}$ and Quan Nguyen$^{1}$% <-this % stops a space
\thanks{This work was supported by the Office of Naval Research grant number
N00014-17-1-2062.}% <-this % stops a space
\thanks{$^{1}$These authors are with the Department of Aerospace and Mechanical Engineering, University of Southern California, Los Angeles, CA 90089
        {\tt\small {\{markherm,luhar,quann\}}@usc.edu}}%
\thanks{$^{2}$This author is recent graduate from the Department of Mechanical Engineering, University of Michigan, Ann Arbor, MI 48109 {\tt\small mclaughlintay16@gmail.com}}%
}

\begin{document}

\maketitle
\thispagestyle{empty}
\pagestyle{empty}

%%%%%%%%%%%%%%%%%%%%%%%%%%%%%%%%%%%%%%%%%%%%%%%%%%%%%%%%%%%%%%%%%%%%%%%%%%%%%%%%
\begin{abstract}
This paper presents a novel omnidirectional gait design and feedback control of a radially symmetric tripedal friction-driven robot. The robot features 3 servo motors mounted on a 3-D printed chassis 7 cm from the center of mass and separated 120 degrees. These motors drive limbs, which impart frictional reactive forces on the body. We first introduce a mathematical model for the robot motion, then show experimental observations performed on a uniform friction surface, which validated the accuracy of the model. This model was then used to create an omnidirectional gait that allows the robot to translate in any direction. %Tripedal symmetry allowed estimation of normal forces for friction-force closure. 
% We demonstrated line following using live feedback from an overhead tracking camera. Proportional-Integral error compensation performance was compared to a basic position update procedure on a rectangular course. 
Based on this gait, we also introduce a Proportional-Integral (PI) feedback control framework that enables the robot to closely follow a desired path. Contrasting with feedforward motion generation, the proposed feedback controller reduced the tracking error by approximately $46\%$. We have successfully demonstrated the approach in our robot hardware for the problem of line following using live feedback from an overhead tracking camera. Our controller is also able to correct for aerodynamic disturbances generated by a high-volume industrial fan with a mean flow speed of $5.5ms^{-1}$, reducing path error by $65\%$ relative to the basic position update procedure. 
\end{abstract}

%%%%%%%%%%%%%%%%%%%%%%%%%%%%%%%%%%%%%%%%%%%%%%%%%%%%%%%%%%%%%%%%%%%%%%%%%%%%%%%%
\section{Introduction}
The vast majority of studies on biological and biomimetic locomotion --- aquatic, terrestrial, or aerial --- have focused on body types exhibiting bilateral symmetry.  This has provided substantial insight into the underlying dynamics, effective gait patterns, and motion control for bilateral actuation. However, few studies have considered locomotion and control for radially symmetric bodies.  Radial symmetries are particularly common in aquatic locomotion. Examples include many members of the phyla Cnidaria (e.g., swimming medusae, or jellyfish; \cite{Cameron1989}) and Echinodermata (e.g., crawling organisms such as sea stars, brittle stars, and sea urchins; \cite{Manuel2009,Astley2012}).  The goal of this paper is to provide physical insight into the crawling locomotion of organisms with radially-symmetric body plans and to demonstrate controlled locomotion using a simple friction-driven tripedal robot.  In particular, we focus on crawling driven by large-scale limb movement. Further, we limit our consideration to locomotion driven by frictional contact with the substrate rather than locomotion enabled by reversible attachment. In this paper, we introduce a novel omnidirectional gait for path following with vision-based feedback control. We demonstrate the effectiveness of our model with a physical robot placed in a turbulent wind field. Our novel normal force estimation procedure, cycle-based gait strategy, and disturbance rejection feedback control method can be useful to other radially symmetric robots seeking translation control in the presence of unmodeled forces.

\begin{figure} [t!]
    \centering
    \includegraphics{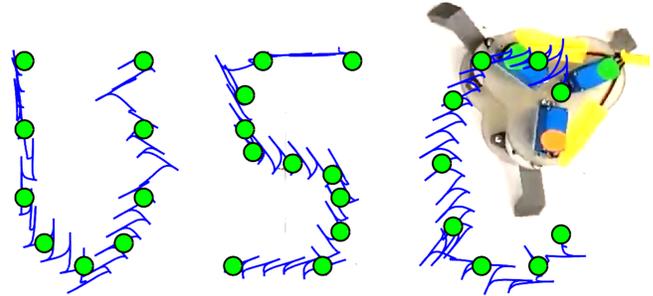}
    \caption{Minimally-actuated tripedal robot demonstrating curve following capabilities. Experimental video: \protect\url{https://youtu.be/F9UxznYtJGM}.}
    \label{fig:experimentTracking}
\end{figure}

Some advantages that radially-symmetric bodies have over bilaterally-symmetric bodies include the ability to change direction quickly, a potential redundancy in the number of appendages driving locomotion, and, generally, an increase in contact points for resistance to exogenous forcing \cite{Arshavskii1976}.  For example, it is well-known that the \textit{lead} limb can be arbitrarily assigned in brittle stars \cite{Cole1913,Astley2012}.  This means that the motion strategy can be modified more easily in case of limb detachment, which is often an escape strategy \cite{Kano2018,Skold1996,Carnevali2006,Matsuzaka2017}.  Previous observations also suggest that changes in crawling direction are achieved by simply defining a new lead limb, and altering the motion of the remaining limbs to follow \cite{Astley2012}.  No rotation of the body is necessary.  Thus, radial symmetry, in conjunction with an even distribution of sensory and actuation systems across all limbs \cite{Lawrence1987}, ensures that there is no \textit{preferred} locomotion direction relative to body orientation.  Finally, having multiple radially-distributed limbs also creates more contact points for adhesion, which becomes important in the case of resisting exogenous forces like fluid forcing.

Some previous studies have also developed radially symmetric bioinspired robots to investigate these body plans. A pentaradial five degree-of-freedom (DOF) robot was recently shown to adapt to limb amputation and retain the ability to translate in a specified direction \cite{Kano2018}. In an earlier study, a pentaradial robot with six actuators in each limb was developed \cite{Lal2008}.  Genetic algorithms were then used to identify optimal gait sequences for this robot from block simulations of a brittle-star model in a physics engine.  This effort showed that neither bilateral nor radial symmetries were locally optimal for translation.  Instead, a complex writhing motion was shown to be most effective. Finally, tripedal robots with wheels attached, otherwise known as \textit{trident snake robots}, have been studied for path planning, geometric control, and optimal control \cite{ishikawa2009development,ishikawa2009tracking,jakubiak2010motion,yamaguchi2012dynamical}. Other researchers have focused on the development of \textit{soft} radially-symmetric robots employing pneumatic actuation \cite{Drotman2017}, shape-memory-alloy actuation \cite{Mao2014}, and electric field manipulation \cite{Otake2002} to yield greater geometric adaptability in complex, unstructured environments.

%\color{red}
%We should also have one paragraph to talk about what are we doing diffidently from prior work and limitation of prior work that inspires us to work on this project. We then can also summary the main contribution of the paper. Sth like
%\color{black}
%The main contributions of this paper with respect to prior work are as follows:
%\begin{itemize}
%    \item The robot is minimally actuated with only three servomotors
%    \item The robot can translate in any direction independent of state
%    \item The robot gait map is independent of friction coefficient
%    \item This gait can facilitate curve following
%\end{itemize}
%(Also emphasize some impressive number here, similar to the last two sentences in the abstract)
% In this paper, we develop and test a minimally-actuated friction driven tripedal robot (Fig.~\ref{fig:experimentTracking}).  We develop a controller that allows this robot to translate in any direction, independent of the current state.  We also show that the resulting gait map is not very sensitive to the friction coefficient at the contact points.  We demonstrate effective curve following in experiments.  Further, we implement an error compensator that reduces path error substantially relative to a basic position update procedure and makes locomotion robust to aerodynamic disturbances. 

The main contributions of the paper are listed below. 
\begin{itemize}
    \item We develop a minimally-actuated friction driven tripedal robot (Fig.~\ref{fig:experimentTracking}).
    \item We present a mathematical model for the robot motion and verify the model in experiment.
    \item We introduce a novel omnidirectional gait that allows the robot to transition in any direction.
    \item We propose a feedback control framework based on PI control to improve tracking performance and compensate for external disturbances.
    \item We pursue laboratory experiments to validate the mathematical model and show the efficacy of our proposed omnidirectional gait, which has an average deviation of $5\%$ for all directions. 
    \item We show successfully path-following capability in the physical robot: the proposed feedback control reduces tracking error for a rectangular trajectory by $46\%$ compared to experiments without PI control. 
    \item We show that our approach is robust to external disturbances: the robot follows a rectangular trajectory in a turbulent wind flow environment (noisy force disturbances) using PI feedback control.
\end{itemize}
\section{Robot and Dynamical Model}

In this section, we describe the robot design including measured physical parameters (e.g. mass, friction coefficient) (\ref{sec:robot}), our mathematical model that we use to describe the motion of the robot (\ref{sec:model}), and compare the results of our simulations with experiments (\ref{sec:modelValid}).

\subsection{Robot Design and Characterization} \label{sec:robot}

% \begin{figure}
%     \centering
%     \includegraphics{figures/RAL_robotSpecs.png}
%     \caption{Physical tripedal system, where servo-motors drive 3-D printed limbs in contact with a surface.}
%     \label{fig:robotSpecs}
% \end{figure}

The tripedal robot, shown in Fig. \ref{fig:experimentTracking}, and Fig \ref{fig:modelDetails}, was designed to approximate the size and geometry of natural sea stars \cite{Hayne2013}.  The robot has limbs of length $l = 7.5$cm evenly distributed around a central structure, attached at an effective radius of $R = 5$cm. Three Hitec HS-5646 submersible servo motors are mounted onto a photopolymer UV-cured support frame to actuate the limbs.  The specific actuation patterns tested are discussed below.  The limbs were 3D printed and designed to minimize material so that the inertial contribution of the limbs ($88$g) could be neglected when compared to the more massive central structure ($800$g).  The total robot mass is $M = 888$g.  The power source and Arduino-based control system were stored externally and connected to the robot via a central cable to reduce mass and ease waterproofing.

\begin{figure}
    \centering
    \includegraphics{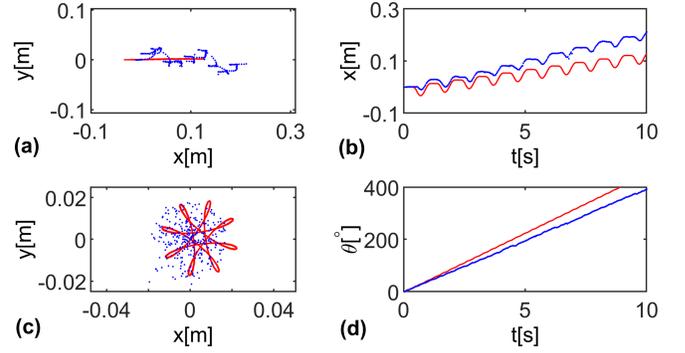}
      \caption{Predicted and measured paths $\mathbf{x}(t)$ for the sandpaper-paper contact for (a) translation gait and (c) rotational gait. For translation, we show the state $x$ vs $t$ in (b), and for rotation, we show state $\xi$ vs $t$ in (d) as these are the states of interest. Experiment paths are traced with blue markers, and simulation paths are traced with red lines.}
    \label{fig:transAndRot}
\end{figure}

Disposable tape strips were attached to the contact points to distribute the normal forces and tune friction. For the model validation experiments shown in Fig.~\ref{fig:transAndRot}, the contact points were covered with 600-grit sandpaper.  Additional validation tests (not shown here) were also carried out with a lower friction polymer tape contacts. The experiments described in this letter were conducted with the robot moving on a sheet of white drafting paper (to ease motion tracking) layered onto a $6.35$mm-thick leveling mat that was stretched across an optical table. The kinetic friction coefficients, $\mu$, for the polymer-paper and sandpaper-paper contacts were measured by releasing a known mass, $m$, from rest on a slope with angle $\beta$, and measuring the velocity achieved by the mass, $V$, after traveling a distance $\Delta x$ down the ramp. Frictional energy loss was estimated from these measurements using the relation
\begin{equation}
\Delta E_f = m g \Delta x \sin{\beta} - \frac{1}{2}mV^2,
\end{equation}
in which the first term on the right-hand side is the change in gravitational potential energy for the sliding mass, and the second term is the gain in kinetic energy.  Assuming that the frictional loss can be modeled as $E_f = \mu N \Delta x$ where $N = m g \cos \beta$ is the normal reaction force acting on the mass, the kinetic friction coefficient can be estimated via the relation
\begin{equation}
\mu = \frac{\Delta E_f}{\Delta x m g \cos \beta}.
\end{equation}

For the sandpaper-paper contact, the friction coefficient was estimated to be $\mu = 0.85 \pm 0.043$. For the polymer-paper contact, the coefficient of friction was estimated to be $\mu = 0.33 \pm 0.012$. 
% For each contact type, the friction coefficient was estimated in nine independent trials, with the mass and inclination angle varying systematically over the ranges $m \in [100g,300g]$ and $\beta \in [15^\circ,55^\circ]$. The $m-\beta$ combinations were chosen to ensure that the uncertainty in the estimated $\mu$ remained within acceptable bounds.  For brevity, details regarding the multivariate uncertainty estimation procedure are not reproduced here. For the polymer-paper friction contact, the coefficient of friction was estimated to be $\mu = 0.33 \pm 0.012$.  For the sandpaper-paper contact, the friction coefficient was estimated to be $\mu = 0.85 \pm 0.043$.

To ensure that the observed robot motion was purely due to limb actuation, we made sure that the dynamic effects due to imperfections in table leveling and tension in the cable tether were negligible. An electronic leveling tool showed that the optical support table where the robot motion was recorded had a tilt no greater than $0.2^\circ$.  For a maximum tilt of $0.2^\circ$, the in-plane force due to gravity is 
\begin{equation}
F_{tilt} = M g \sin(0.2\pi/180) \approx 0.0035 Mg
\end{equation}
Assuming that the normal force at each of the three contact points is on average $Mg/3$, the friction force at each contact point for the sandpaper-paper case is expected to be
\begin{equation}
F_f \approx \mu \frac{Mg}{3} \approx 0.28 Mg.
\end{equation}
 (For clarity, this equation is not used for modeling purposes, but rather to show the effect of small angle tilts are negligible to the dynamics.) In other words, for this high-friction case, the in-plane friction force at each contact is expected to be roughly two orders of magnitude larger than the total force due to tilting.  Similarly, cable tension was measured using a hanging scale and found to be negligible compared to the total weight of the system.  Thus, cable tension is not expected to play a dynamic role either.

\subsection{Physics-Based Model}\label{sec:model}

\begin{figure}
    \centering
    \includegraphics{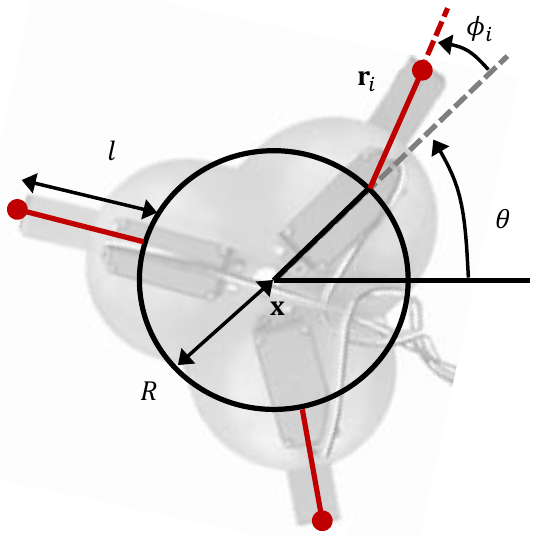}
    \caption{Simplified representation of tripedal system. $\mathbf{x}$ is the position coordinate of the center of mass with respect to a fixed reference frame, $\xi$ is the rotation of the body based on the hinge point of a specified limb, and $\phi_i$ is the local limb rotation.}
    \label{fig:modelDetails}
\end{figure}

The mathematical model is based on conservation laws for linear and angular momentum with a nonlinear frictional forcing.  The model assumes that the translational inertia and rotational inertia for the limbs are negligible relative to those for the more massive central structure.  The friction forces are assumed to act at the point of contact at the end of each robot limb (see Fig.~\ref{fig:modelDetails}).  

To model the friction force, we use a simplified version of the Coulomb friction law in which no distinction is made between static and kinetic friction \cite{olsson1998friction,pennestri2016review}. Specifically, the friction force, $\mathbf{F}_i = [F_{i,x},F_{i,y}]$, acting at the contact point at the end of each limb in the $x-y$ plane, $\mathbf{r}_i = [r_{i,x},r_{i,y}]$, is modeled as
\begin{equation}\label{friction}
\mathbf{F}_i = - \mu N_i \frac{\dot{\mathbf{r}}_i}{|\dot{\mathbf{r}}_i|}.
\end{equation}
Here, $N_i$ is the normal force at contact point $i$, $\mu$ is the measured kinetic friction coefficient, and $\dot{\mathbf{r}}_i = [\dot{r}_{i,x},\dot{r}_{i,y}]$ is the velocity of the contact point at the end of each limb.  Note that $\mathbf{r}_i$ and $\dot{\mathbf{r}}_i$ can be expressed in terms of the robot state vector, $\mathbf{z}= [\mathbf{x}, \dot{\mathbf{x}}, \xi, \dot{\phi}]$, the actuation angles and rotation rates, $\phi_i$ and $\dot{\phi_i}$, and the geometric constants, $R$ and $l$, using simple trigonometric relations (see Fig.~\ref{fig:modelDetails}).
Normal forces are computed by combining a vertical force balance ($N_1 + N_2 + N_3 = M g$) with torque balances about the robot center of mass:
\begin{equation} \label{normalforces}
\left[\begin{array}{c c c}
1 & 1 & 1 \\ 
r_{1,x} - x &	r_{2,x} - x & 	r_{3,x} - x \\
r_{1,y} - y &	r_{2,y} - y & 	r_{3,y} - y 
\end{array}\right]	
\left[\begin{array}{c }
N_1\\ N_2 \\N_3 
\end{array}\right]	=
\left[\begin{array}{c }
M g \\ 0 \\ 0 
\end{array}\right].
\end{equation} 
The second and third lines in the equation above ensure that there are no net torques about the robot center-of-mass due to the normal forces.
	
Under the assumptions stated above, conservation laws for linear momentum in the horizontal plane can be expressed as
\begin{equation}\label{Mom1}
M \ddot{x} = \sum \limits_{i=1}^{3} F_{i,x} 
\end{equation}
and
\begin{equation}\label{Mom2}
M \ddot{y} = \sum \limits_{i=1}^{3} F_{i,y}, 
\end{equation} 
respectively.
The conservation law for angular momentum can be expressed compactly as
\begin{equation}\label{Mom3}
J \ddot{\xi} = \sum \limits_{i=1}^{3}  (\mathbf{r_i} - \mathbf{x}) \times \mathbf{F}_i,
\end{equation}
where $J$ represents the rotational inertia, which is estimated from the mass distribution of the central support structure.  

The system of ordinary differential equations shown in (\ref{Mom1}-\ref{Mom3}) is solved numerically to yield predictions for robot translation ($\mathbf{x}$) and rotation ($\xi$) using the MATLAB \textsf{ode45} algorithm for an adaptive time-step 4th-order Runge-Kutta solver.  Recall that the location of the contact points relative to the robot center, $\mathbf{r}_i - \mathbf{x}$, depends on the limb angles, $\phi_i(t)$ (Fig.~\ref{fig:modelDetails}).  Therefore, the prescribed actuation angles $\phi_i(t)$ appear directly in all three conservation laws via the friction terms dependent on $\mathbf{r}_i$ and $\dot{\mathbf{r}}_i$.  Also, keep in mind that these model simulations involve no \textit{tuning} parameters.  All geometric and dynamic variables appearing in the governing equations (e.g., $M, J, l, R, \mu$) are obtained from independent measurements.

\subsection{Model Validation} \label{sec:modelValid}

Model validation tests demonstrated that the robot can rotate and translate independently. These motions can be achieved with sinusoidal actuation at the limbs. If two limbs are in anti-phase, and the third limb is motionless, the robot translates in the direction of the inactive limb, as shown in Fig.~\ref{fig:transAndRot}(a),(b).  Here, the active limbs were prescribed to move sinusoidally with amplitude 30$^\circ$ and frequency $f = 1$ Hz. If all three limbs are actuated with the same amplitude and frequency, but with a phase shift of $120^\circ$ relative to one another, the robot rotates in place as shown in Fig.~\ref{fig:transAndRot}(c),(d).  Note that these experiments also confirm that the mathematical model adequately reproduces the dynamics of the physical system for the high-friction contact. Though experimental results agree well with simulation predictions, integrated errors from unmodeled disturbances do become significant without feedback to correct trajectories. Some sources of uncertainty include model simplifications and physical system imperfections (uneven mass distribution, surface irregularities). We also observed close agreement between model predictions and experimental results for the low-friction polymer-paper contract surface with $\mu=0.33 \pm 0.012$ (not shown here). This modeling framework is used for gait and control design in the following section.

\section{Gait and Control Design}
 In this section, we discuss the development and parametrization of the omnidirectional gait (\ref{gait1} and \ref{gait2}) before discussing controller design (\ref{Cntrl}). 

\subsection{Omnidirectional Gait} \label{gait1}

The predictive model showed that an omnidirectional gait would allow for immediate translation in any direction. This new gait was a variation of the previously introduced translational gait. As shown in Fig. \ref{fig:transAndRot}(a,b), when two limbs operate in anti-phase to one another, and the third limb is inactive, the robot translates in the direction of the inactive limb. However, simulations show that sinusoidal actuation of the inactive limb produces translation at a non-zero angle relative to the previously inactive limb. This angle of translation, $\theta_D$ varied depending on the amplitude, $\alpha$ of the previously inactive limb's sinusoidal motion. In this new gait, the previously inactive limb is defined as the $\alpha$-limb because the amplitude of its sinusoidal motion is $\alpha$. The other two limbs operate in anti-phase to one another, with constant amplitude sinusoidal motion. Simulations generated (details in Section \ref{gait2}) a map of sinusoid amplitude to angle of translation, $\mathbf{M}$ (Fig.~\ref{fig:MappingPlot}).

Importantly, because of the radial symmetry of the model, any limb can be made the $\alpha$-limb, allowing the model to translate in any direction from any initial configuration. The desired direction of motion ($\theta_D \in [0,360]^\circ$) is divided into six zones spanning $60^\circ$ of translation angle, as shown in Table \ref{tab:Zones}. For $\theta_D \in [0,60]^\circ$, lead limb amplitude $\alpha$ is calculated using a mapping of $\theta_D$ to $\mathbf{a} = [a_1, a_2, a_3]^\circ = [\alpha, 30, -30]^\circ$, in which $a_i$ represents the oscillation amplitude for limb $i$ and $a_1$ is the lead (or $\alpha$) limb. The mapping between $\theta_D \in [0,60]^\circ$ and $\alpha$ is plotted in Fig. \ref{fig:MappingPlot}. This function is combined with different permutations of limb amplitudes to span $\theta_D \in [0,360]^\circ$.

The outputs of the MATLAB block in Fig. \ref{fig:MATLABDiagram} are the gait parameters $\mathbf{a}=[a_{1}, a_{2}, a_{3}]$ defining the sinusoidal motion of each limb:
\begin{equation}
    \phi_i = a_i \sin(2 \pi f t).
\end{equation}
In the expression above, $f$ is the frequency of limb oscillation, which was set to 1 Hz in both the simulations and the experiments. The limb angle $\phi_i$ can be visualized in Fig.~\ref{fig:modelDetails}. One of the gait parameters, $[a_{1}, a_{2}, a_{3}]$, is set to equal $\alpha$. The value of $\theta_D$ is used to determine what index of $\mathbf{a}$ is set to the $\alpha$-value, following the structure shown in Table \ref{tab:Zones}.

\begin{comment}
This is discussed further below in the context of controller design. This framework, coupled with feedback control, was then used to demonstrate path following capabilities in simulations and experiments. Both the robot and simulation experiments update the position gait parameters every cycle.

\end{comment}

\begin{comment}

\begin{figure}
    \centering
    \includegraphics{figures/RAL_robotExperiment.png}
      \caption{Schematic showing how tracking information from PixyCam2 is used as feedback in MATLAB to update the gait pattern.}
    \label{fig:experimentSetup}
\end{figure}

\begin{figure}
   \includegraphics[trim={2 376 713 10},clip]{blockDiagram.pdf}
   \caption{A schematic showing the feedback processing and gait selection in the MATLAB block in Fig.~\ref{fig:experimentSetup}.}
   \label{fig:MATLABDiagram}
\end{figure}

\end{comment}

\subsection{Gait Map} \label{gait2}
\begin{figure}
    %\centering
    \includegraphics{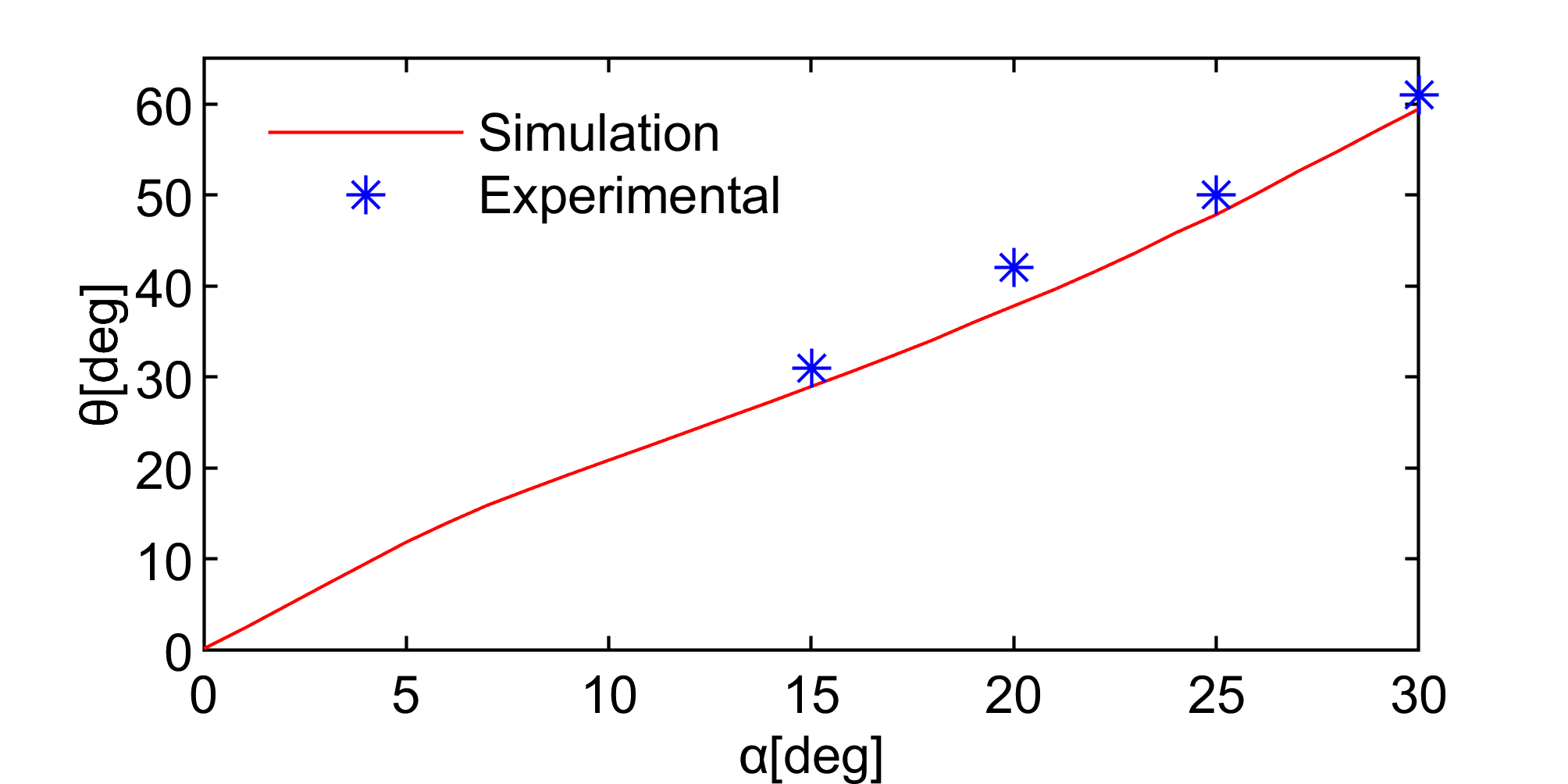}
      \caption{A comparison of the simulated average map, $\mathbf{M}_{avg}$ to experimentally collected data points. This map is used to prescribe an $\alpha$-limb gait parameter, $\alpha$, given a desired angle of translation, $\theta$.}
    \label{fig:MappingPlot}
\end{figure}

Open loop simulations with $\alpha \in [0,30] ^\circ$ and with non-$\alpha$ limb amplitudes $30^\circ$ generated $\mathbf{M}$ (Fig. \ref{fig:MappingPlot}). A maximum value of $\alpha = 30^\circ$ was selected due to limitations of the physical model. Over the course of a trial, $\alpha$ was held constant, and the resulting translation angle $\theta$ was measured relative to the model reference frame for each cycle of motion. The $\theta$ values of each cycle were then averaged over the entire trial. This averaging produced the average angle of translation, $\theta_{avg}$. There was minimal variation in $\theta$ after the first two cycles of motion.

\begin{table}
\caption{%The range of input $\theta$ values to the Gait selection block, and the corresponding form of the output gait parameter vector, $\mathbf{a}$.Here, $\theta$ is measured relative to the frame of reference of the robot. See Section \textbf{\nameref{Cntrl}} for and explanation of the structure of $\mathbf{a}$. 
Relationship between desired translation direction and the gait parameters for each limb.  All angles are prescribed in degrees.}
    \centering
   \begin{tabular}{c c c}
\hline
\textbf{Zone} & Desired Translation Angle ($\theta_D$) & Gait Parameters ($\mathbf{a}$)  \\
\hline
1 & $0 <\theta \leq 60$ & [$\alpha$, 30, $-30$]\\
2 & $60 <\theta \leq 120$ & [$-30$, $-\alpha$, 30]\\
3 & $120 <\theta \leq 180$ & [$-30$, $\alpha$, 30]\\
4 & $180 <\theta \leq 240$ & [30, $-30$, $-\alpha$]\\
5 & $240 <\theta \leq 300$ & [30, $-30$, $\alpha$]\\
6 & $300 <\theta \leq 360$ & [$-\alpha$, 30, $-30$]\\
\hline
\end{tabular}
    \label{tab:Zones}
\end{table}

This procedure was repeated in three different different friction environments: $\mu=[0.33, 0.59, 0.87]$. These values were selected based upon the friction coefficient values tested with the physical robot. The differences in the mappings generated from these three friction coefficients was small. As a result, we assumed a universal mapping, $\mathbf{M}_{avg}$, to provide sufficient accuracy independent of friction magnitude. Fig.~\ref{fig:MappingPlot} shows this averaged map obtained using the predictive model (red curve), together with $[\theta,\alpha]$ measurements made in experiments using the tripedal robot (blue symbol).

%%More info than necessary
%Map insensitivity to friction was further evaluated using the predictive model. During this study, the model was run through a randomly generated course, in three different friction environments, described by $\mu=[0.33, 0.59, 0.85]$. The model completed the course, in each environment, using all four maps, $\mathbf{M}_{0.33}$, $\mathbf{M}_{0.59}$, $\mathbf{M}_{0.85}$, and $\mathbf{M}_{avg}$.  Using total time to complete the course as a performance metric, $T_{c}$, there was $<1\%$ variation amongst all four maps. As a result, $\mathbf{M}_{avg}$ was used in all simulations and physical robot experiments, regardless of the friction environment.

 \subsection{Controller Design}
 \label{Cntrl}

 \begin{figure*}[h!t]
   \centering    
   \includegraphics[trim={2 165 485 3},clip]{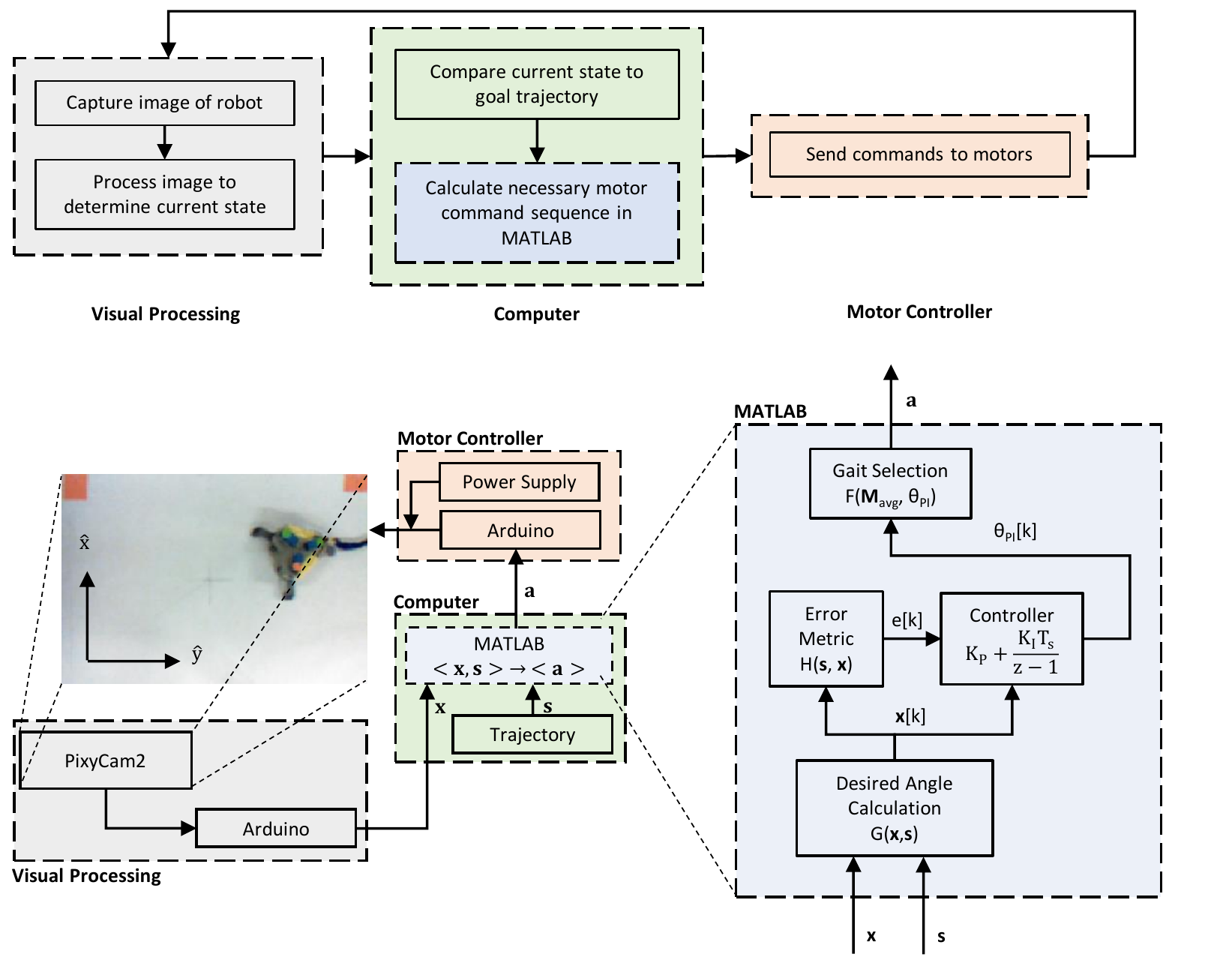}
   \caption{Flow chart (top) and diagram (bottom) showcasing feedback processing and gait selection.}
   \label{fig:MATLABDiagram}
\end{figure*}

Fig. \ref{fig:MATLABDiagram} provides an overview of the feedback tracking procedure implemented in the experiments. The position vector $\mathbf{x}=[x,y,\xi]$ and trajectory $\mathbf{s}$ are inputs to the gait mapping implemented in MATLAB. The inputs to MATLAB, $\mathbf{x}$ and $\mathbf{s}$, are used to determine the desired angle of translation, $\theta_D = G(\mathbf{x},\mathbf{s})$, using the arctangent function. A simple PI controller, shown in the blue MATLAB block of Fig. \ref{fig:MATLABDiagram}, was implemented to compensate for steady-state error. The PI controller outputs an adjusted desired angle of translation, $\theta_{PI}$ according the equation,
\begin{equation}
    \theta_{PI}[k] = (K_p + \frac{K_I T_s}{z-1})e[k] + \theta_D[k]
\end{equation}
where $k$ is the discrete time variable, $e$ is the error signal, $K_P$ is the proportional gain, $K_I$ is the integral gain, $T_s$ is the duration in seconds, and $z$ is the z-transform operator. An error metric, $H(\mathbf{s},\mathbf{x})$, estimates $e$ by calculating the minimum distance from a line connecting target points and the position measurement. The Gait Selection block uses $\theta_{PI}$, along with the average gait map, $\mathbf{M}_{avg}$ to select $\alpha$ and send amplitudes, $\mathbf{a}$, to the motors. Depending on $\theta_{PI}$, one of the six zones listed in Table~\ref{tab:Zones} is selected and the mapping $\mathbf{M}_{avg}$ is used to calculate the value of $\alpha$ and the gait parameter vector $\mathbf{a}$.

\begin{figure*}
    \centering
    \includegraphics{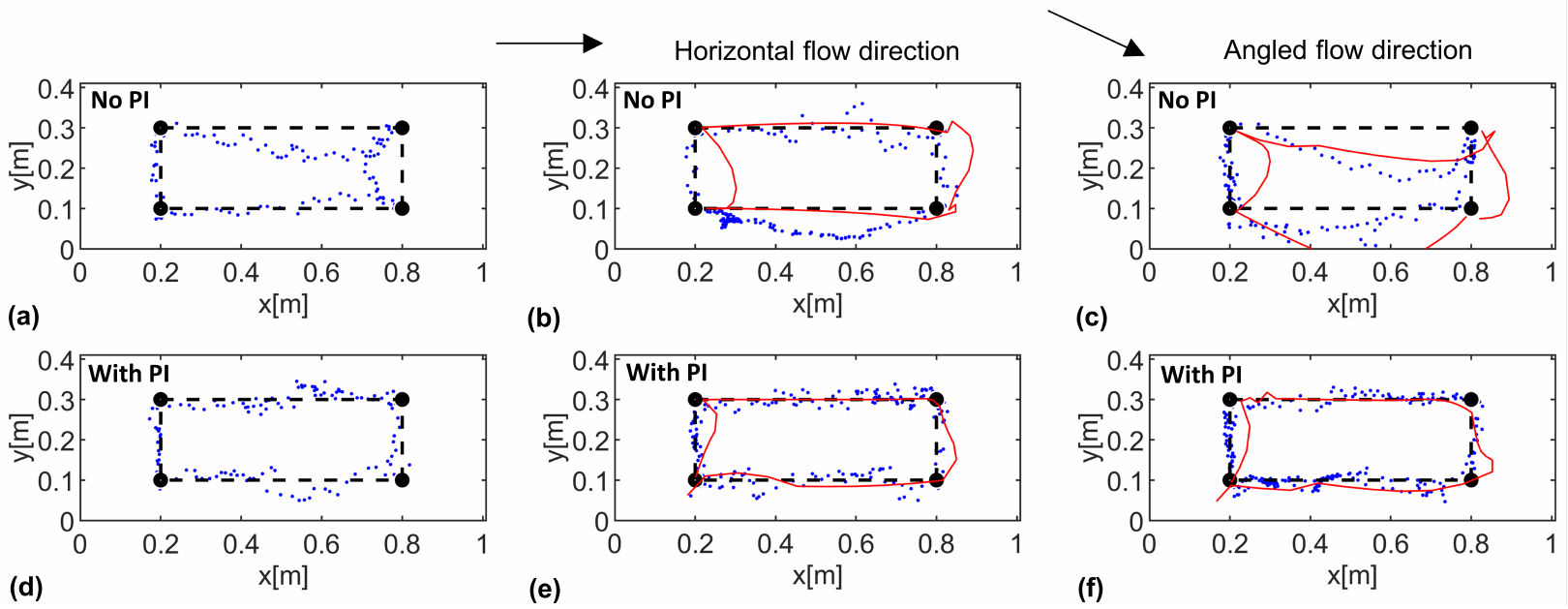}
    \caption{Four points are used for the robot to trace a square path in the presence of a variable wind flow field where (a)-(c) are without error compensation and (d)-(f) are with PI error compensation. (a),(d) Plots are without wind flow, (b),(e) plots are with 0$^\circ$ wind flow, and (c),(f) plots are wind flow at 25$^\circ$ with respect to the x-direction. Blue markers show experimental measurements while red curves show simulation results.}
    \label{fig:controlWithFlow}
\end{figure*}

\section{Path Following Demonstration}
To illustrate path following capabilities in simulation and experiments, a rectangular path specified by 4 target points was used (see Fig.~\ref{fig:controlWithFlow}).
%The robot followed a rectangular path specified by 4 target points, shown in Fig.~\ref{fig:controlWithFlow}. 
In the physical experiments, we used a surface with an unknown friction coefficient between the low-friction polymer-paper case ($\mu = 0.33$) and the high-friction sandpaper-paper case ($\mu = 0.88$) to demonstrate the friction-independence of $\mathbf{M_{avg}}$.  In addition, an industrial floor fan was used to generate wind flow across the test surface to study the effect of aerodynamic disturbances on path following capabilities. A Protmex 6252A handheld anemometer measured wind speed at both ends of the testing platform. The edge located closest to the fan had a flow speed of $6.4 \pm 0.2 $ms$^{-1}$ and the edge furthest from the fan had a flow speed of $4.5 \pm 0.2 $ms$^{-1}$. Thus, there were significant velocity gradients across the surface, making for a complex flow field. 

In the numerical simulations, aerodynamic drag was modeled as $F_d = (1/2) \rho C_d A_d v^2$ where $\rho$ is the density of air, $C_d$ is a drag coefficient, $A_d$ is the frontal area of the robot, and $v$ is wind speed. The drag coefficient was assumed to be $C_d =1$, corresponding roughly to the drag coefficient of a cubic bluff body.  The frontal area for the robot was estimated to be $A_d = 0.02$m$^2$.  The wind speed was set at $v=5.5$ms$^{-1}$, an average of the velocities measured at each end of the test section. We note that these values were for simulation purposes only to provide a reference for the experiment results. The controller does not use the fluid dynamics of the system for gait selection. The flow is considered to be a general disturbance. This model may differ from the physical system where we assume the flow field is assumed uniform and constant, there are no ground effects, the drag coefficient does not change with position, and there are no lift effects.

Proportional gain was $K_P = 15$ deg cm$^{-1}$ for rapid correction to path deviations, and integral gain was $K_I = 1$ deg cm$^{-1}$ to reduce steady-state error. We selected these values based on experimental tuning with an update rate of $f = 1$ Hz (i.e., every actuation cycle for the limbs). From Fig.~\ref{fig:controlWithFlow}(a) it can be seen that, without proportional-integral (PI) error compensation, the robot does not follow a straight line. This is likely a result of error from unmodeled disturbances in $\mathbf{M_{avg}}$, imperfections in robot manufacturing, error in gait execution, imbalance from cable tether, and experiment surface inhomogeneity. Fig.~\ref{fig:controlWithFlow}(d) shows that the PI controller corrects for this unmodeled drift. Fig.~\ref{fig:controlWithFlow}(b),(c) show the effect of wind in the environment on path following without PI error compensation. The robot is particularly sensitive to wind speed angled at 25$^\circ$ with respect to the x-direction, where deviations from the nominal trajectory are amplified. The PI controller successfully compensates for this in Fig.~\ref{fig:controlWithFlow}(f).  Note that the simulation results shown Fig.~\ref{fig:controlWithFlow} are in reasonable qualitative agreement with the experimental tracks for the cases with background flow.

Table~\ref{tab:QMetrics} shows two performance metrics for the tracking experiments: cumulative path following error $\Delta = \sum |e|$ (m), where $e$ is the perpendicular distance to the nominal path, and the completion time $T_{c}$ (s).   
%Performance metrics, defined as $Q_1 = \sum | \text{error} |$ (m), where error is the perpendicular distance to the nominal path, and $Q_2 =$ time to complete course (s), are assigned to assess the effect of the controller implementation and results are compiled in table \ref{tab:QMetrics}. 
In all flow environments, $\Delta$ is reduced by implementation of the PI controller.  For example, $\Delta$ is reduced by over 65\% for the cases with flow in the $x$-direction. However, $T_c$ increases with added control for both the no flow and angled flow cases. This is because the robot spends time correcting its trajectory rather than going directly to the way points. A time-optimal path in the presence of flow is not straightforward, and is a topic of further investigation.

\begin{table}
    \centering
    \caption{Experimental performance in terms of cumulative path following error ($\Delta$, m) and completion time ($T_c$, s).  Rows and columns correspond to Fig.~\ref{fig:controlWithFlow}.}
    \begin{tabular}{c c  |c c  |c c  |c c}
        & & \multicolumn{2}{|c|}{No flow} &\multicolumn{2}{|c|}{$0^\circ$ flow} & \multicolumn{2}{|c}{$25^\circ$ flow}\\
    & & $\Delta$ & $T_c$ & $\Delta$ & $T_c$ & $\Delta$ & $T_c$\\
    \hline
    \multicolumn{2}{c|}{No PI} & 3.39 & 16 & 8.12 & 39 & 5.34 & 21\\
    \multicolumn{2}{c|}{With PI} & 1.81 & 18 & 2.81 & 26 & 2.73 & 36\\
    \end{tabular}
    \label{tab:QMetrics}
\end{table}

Another demonstration of robot path following capability is provided in Fig.~\ref{fig:experimentTracking}, which shows that complex curved paths can be tracked by placing intermediate target points. For this figure, we show tracking with no error compensation using the standard $\mathbf{M}_{avg}$ map.

\section{Conclusion}
We have shown that a momentum-conservation model can accurately reproduce motion for a friction driven tripedal robot. Normal forces are accurately predicted by solving an inverse problem from a  force/torque balance system of equations. This principle can be extended to other three-point systems where normal forces are of interest. This model facilitated generation of a gait-map which predicted that the robot can translate in any direction independent of the current state. Open loop experiments confirmed this, and, with this mapping, the physical system demonstrated effective path following with an unknown friction coefficient and in a high-speed wind flow environment. The empirical gait-map control strategy illustrated in this work can be applied to other nonlinear, non-holonomic systems where analytical approaches may not be possible. This study shows how a minimally actuated, radially symmetric robot can achieve path following by exploiting asymmetrical gait patterns.  

Future work will target time optimal path following in the presence of background flows, fluid dynamic characterization of the physical robot, and locomotion on heterogeneous surfaces.

%\addtolength{\textheight}{-12cm}   
% This command serves to balance the column lengths on the last page of the document manually. It shortens the textheight of the last page by a suitable amount.
% This command does not take effect until the next page so it should come on the page before the last. Make sure that you do not shorten the textheight too much.

%%%%%%%%%%%%%%%%%%%%%%%%%%%%%%%%%%%%%%%%%%%%%%%%%%%%%%%%%%%%%%%%%%%%%%%%%%%%%%%%

%%%%%%%%%%%%%%%%%%%%%%%%%%%%%%%%%%%%%%%%%%%%%%%%%%%%%%%%%%%%%%%%%%%%%%%%%%%%%%%%

%%%%%%%%%%%%%%%%%%%%%%%%%%%%%%%%%%%%%%%%%%%%%%%%%%%%%%%%%%%%%%%%%%%%%%%%%%%%%%%%
%\section*{APPENDIX}
%Appendixes should appear before the acknowledgment.

%\section*{ACKNOWLEDGMENT}
%The preferred spelling of the word ÒacknowledgmentÓ in America is without an ÒeÓ after the ÒgÓ. Avoid the stilted expression, ÒOne of us (R. B. G.) thanks . . .Ó  Instead, try ÒR. B. G. thanksÓ. Put sponsor acknowledgments in the unnumbered footnote on the first page.

%%%%%%%%%%%%%%%%%%%%%%%%%%%%%%%%%%%%%%%%%%%%%%%%%%%%%%%%%%%%%%%%%%%%%%%%%%%%%%%%

\bibliographystyle{IEEEtran}
\bibliography{Tripedal}

\begin{thebibliography}{10}
\providecommand{\url}[1]{#1}
\csname url@rmstyle\endcsname
\providecommand{\newblock}{\relax}
\providecommand{\bibinfo}[2]{#2}
\providecommand\BIBentrySTDinterwordspacing{\spaceskip=0pt\relax}
\providecommand\BIBentryALTinterwordstretchfactor{4}
\providecommand\BIBentryALTinterwordspacing{\spaceskip=\fontdimen2\font plus
\BIBentryALTinterwordstretchfactor\fontdimen3\font minus
  \fontdimen4\font\relax}
\providecommand\BIBforeignlanguage[2]{{%
\expandafter\ifx\csname l@#1\endcsname\relax
\typeout{** WARNING: IEEEtran.bst: No hyphenation pattern has been}%
\typeout{** loaded for the language `#1'. Using the pattern for}%
\typeout{** the default language instead.}%
\else
\language=\csname l@#1\endcsname
\fi
#2}}

\bibitem{Cameron1989}
J.~L. Cameron and P.~V. Fankboner, ``Reproductive biology of the commercial sea
  cucumber parastichopus californicus (stimpson)(echinodermata: Holothuroidea).
  ii. observations on the ecology of development, recruitment, and the juvenile
  life stage,'' \emph{Journal of Experimental Marine Biology and Ecology}, vol.
  127, no.~1, pp. 43--67, 1989.

\bibitem{Manuel2009}
M.~Manuel, ``Early evolution of symmetry and polarity in metazoan body plans,''
  \emph{Comptes rendus biologies}, vol. 332, no. 2-3, pp. 184--209, 2009.

\bibitem{Astley2012}
H.~C. Astley, ``Getting around when you’re round: quantitative analysis of
  the locomotion of the blunt-spined brittle star, ophiocoma echinata,''
  \emph{Journal of Experimental Biology}, vol. 215, no.~11, pp. 1923--1929,
  2012.

\bibitem{Arshavskii1976}
Y.~I. Arshavskii, S.~Kashin, N.~Litvinova, G.~Orlovskii, and A.~Fel'dman,
  ``Coordination of arm movement during locomotion in ophiurans,''
  \emph{Neurophysiology}, vol.~8, no.~5, pp. 404--410, 1976.

\bibitem{Cole1913}
L.~J. Cole, ``Direction of locomotion of the starfish (asterias forbesi),''
  \emph{Journal of Experimental Zoology}, vol.~14, no.~1, pp. 1--32, 1913.

\bibitem{Kano2018}
T.~Kano, E.~Sato, T.~Ono, H.~Aonuma, Y.~Matsuzaka, and A.~Ishiguro, ``A brittle
  star-like robot capable of immediately adapting to unexpected physical
  damage,'' \emph{Royal Society open science}, vol.~4, no.~12, p. 171200, 2017.

\bibitem{Skold1996}
M.~Sk{\"o}ld and R.~Rosenberg, ``Arm regeneration frequency in eight species of
  ophiuroidea (echinodermata) from european sea areas,'' \emph{Journal of Sea
  Research}, vol.~35, no.~4, pp. 353--362, 1996.

\bibitem{Carnevali2006}
M.~C. Carnevali, ``Regeneration in echinoderms: repair, regrowth, cloning,''
  \emph{Invertebrate Survival Journal}, vol.~3, no.~1, pp. 64--76, 2006.

\bibitem{Matsuzaka2017}
Y.~Matsuzaka, E.~Sato, T.~Kano, H.~Aonuma, and A.~Ishiguro, ``Non-centralized
  and functionally localized nervous system of ophiuroids: evidence from
  topical anesthetic experiments,'' \emph{Biology open}, vol.~6, no.~4, pp.
  425--438, 2017.

\bibitem{Lawrence1987}
J.~M. Lawrence, \emph{Functional biology of echinoderms}.\hskip 1em plus 0.5em
  minus 0.4em\relax Croom Helm, 1987.

\bibitem{Lal2008}
S.~P. Lal, K.~Yamada, and S.~Endo, ``Evolving motion control for a modular
  robot,'' in \emph{International Conference on Innovative Techniques and
  Applications of Artificial Intelligence}.\hskip 1em plus 0.5em minus
  0.4em\relax Springer, 2007, pp. 245--258.

\bibitem{ishikawa2009development}
M.~Ishikawa, Y.~Minami, and T.~Sugie, ``Development and control experiment of
  the trident snake robot,'' \emph{IEEE/ASME Transactions on Mechatronics},
  vol.~15, no.~1, pp. 9--16, 2009.

\bibitem{ishikawa2009tracking}
M.~Ishikawa, P.~Morin, and C.~Samson, ``Tracking control of the trident snake
  robot with the transverse function approach,'' in \emph{Proceedings of the
  48h IEEE Conference on Decision and Control (CDC) held jointly with 2009 28th
  Chinese Control Conference}.\hskip 1em plus 0.5em minus 0.4em\relax IEEE,
  2009, pp. 4137--4143.

\bibitem{jakubiak2010motion}
J.~Jakubiak, K.~Tcho{\'n}, and M.~Janiak, ``Motion planning of the trident
  snake robot: An endogenous configuration space approach,'' in \emph{ROMANSY
  18 Robot Design, Dynamics and Control}.\hskip 1em plus 0.5em minus
  0.4em\relax Springer, 2010, pp. 159--166.

\bibitem{yamaguchi2012dynamical}
H.~Yamaguchi, ``Dynamical analysis of an undulatory wheeled locomotor: a
  trident steering walker,'' \emph{IFAC Proceedings Volumes}, vol.~45, no.~22,
  pp. 157--164, 2012.

\bibitem{Drotman2017}
D.~Drotman, S.~Jadhav, M.~Karimi, P.~Dezonia, and M.~T. Tolley, ``3d printed
  soft actuators for a legged robot capable of navigating unstructured
  terrain,'' in \emph{2017 IEEE International Conference on Robotics and
  Automation (ICRA)}.\hskip 1em plus 0.5em minus 0.4em\relax IEEE, 2017, pp.
  5532--5538.

\bibitem{Mao2014}
S.~Mao, E.~Dong, H.~Jin, M.~Xu, S.~Zhang, J.~Yang, and K.~H. Low, ``Gait study
  and pattern generation of a starfish-like soft robot with flexible rays
  actuated by smas,'' \emph{Journal of Bionic Engineering}, vol.~11, no.~3, pp.
  400--411, 2014.

\bibitem{Otake2002}
M.~Otake, Y.~Kagami, M.~Inaba, and H.~Inoue, ``Motion design of a
  starfish-shaped gel robot made of electro-active polymer gel,''
  \emph{Robotics and Autonomous Systems}, vol.~40, no. 2-3, pp. 185--191, 2002.

\bibitem{Hayne2013}
K.~J. Hayne and A.~R. Palmer, ``Intertidal sea stars (pisaster ochraceus) alter
  body shape in response to wave action,'' \emph{Journal of Experimental
  Biology}, vol. 216, no.~9, pp. 1717--1725, 2013.

\bibitem{olsson1998friction}
H.~Olsson, K.~J. {\AA}str{\"o}m, C.~C. De~Wit, M.~G{\"a}fvert, and
  P.~Lischinsky, ``Friction models and friction compensation,'' \emph{Eur. J.
  Control}, vol.~4, no.~3, pp. 176--195, 1998.

\bibitem{pennestri2016review}
E.~Pennestr{\`\i}, V.~Rossi, P.~Salvini, and P.~P. Valentini, ``Review and
  comparison of dry friction force models,'' \emph{Nonlinear dynamics},
  vol.~83, no.~4, pp. 1785--1801, 2016.

\end{thebibliography}
% \bibliography{TripedalAbrv,Tripedal}

\end{document}